\documentclass{article}
\usepackage{times}

\usepackage{graphicx, subfigure, natbib, amsmath} 
\usepackage{algorithm, algorithmic, afterpage}
\usepackage{hyperref, enumerate, enumitem}
 
\usepackage{subfigure} 

\usepackage[accepted]{icml2014}
\usepackage{bm, amsmath, verbatim}

\usepackage{tikz}
\usetikzlibrary{fit,positioning}

\icmltitlerunning{Sequential Monte Carlo Bandits}

\graphicspath{{figures/}}

\begin{document} 

\twocolumn[
\icmltitle{Sequential Monte Carlo Bandits}

\icmlauthor{Michael Cherkassky}{mcherkassky@gmail.com}
\icmladdress{Pipewave Inc., Cambridge, MA}
\icmlauthor{Luke Bornn}{bornn@stat.harvard.edu}
\icmladdress{Department of Statistics, Harvard University, Cambridge, MA}

\icmlkeywords{boring formatting information, machine learning, ICML}

\vskip 0.3in
]

\begin{abstract}

In this paper we propose a flexible and efficient framework for handling multi-armed bandits, combining sequential Monte Carlo algorithms with hierarchical Bayesian modeling techniques. The framework naturally encompasses restless bandits, contextual bandits, and other bandit variants under a single inferential model.  Despite the model's generality, we propose efficient Monte Carlo algorithms to make inference scalable, based on recent developments in sequential Monte Carlo methods.  Through two simulation studies, the framework is shown to outperform other empirical methods, while also naturally scaling to more complex problems for which existing approaches can not cope. Additionally, we successfully apply our framework to online video-based advertising recommendation, and show its increased efficacy as compared to current state of the art bandit algorithms. 

\end{abstract} 

\section{Introduction}

The use of multi-armed bandit (MAB) problems to represent sequential decision-making processes has received increasing attention in the literature, with contributions ranging from novel algorithms to strong theoretical results. The fundamental problem which MABs address, that of sequential finite resource allocation, has found application in diverse fields including control theory, advertising, and portfolio allocation. Our movivation arises from a novel dataset in the area of online video advertising, where the content provider seeks to optimize clickthrough rates from advertisements both within and around video content. Given features of the user such as their country and browser, we wish to display the advertisement with the highest probability of the user clicking through.  Further, early investigations suggest that advertising effectiveness changes over time, as such we must account for time-varying clickthrough rates. 

The classic MAB problem, and the origin of its moniker, is to imagine one is sitting in front of a potentially infinite set of slot machines, each with a different expected reward, and at each time one must decide which machine's arm to pull. If one finds an arm that initially performs well, is it better to continue pulling the arm of that machine or exploring to find alternative, potentially higher payout, arms? MAB problems address this issue of balancing \textit{exploitation} of a strategy you know to be best given your current knowledge vs. \textit{exploration} to test alternative strategies.

Some simple, though naive, approaches for MAB problems are the full exploration strategy, which equally allocates actions across all of the arms regardless of past performance, and a Markovian strategy, which replays the current arm if success is achieved and otherwise picks an alternative arm at random \citep{robbins1956sequential}. Both of these strategies ignore the full history of past rewards, and hence ignore relevant information for finding an optimal strategy.  An alternative is to be greedy -- after an initial exploration period choose the arm at each time which has had the highest observed average reward.  Problematically, new arms or those which due to randomness had small initial reward in the exploration period will not be subsequently sampled, leading to possible convergence to a suboptimal strategy.  A variant on the greedy approach is an $\epsilon$-greedy method, which starts with a greedy strategy but with probability $\epsilon$ chooses a different arm at each time.

More probabilistic approaches also exist, such as those which create $100(1-\alpha)\%$ confidence intervals for the expected reward from each arm and select arms based on features of this interval, such as choosing the arm with the highest upper bound \citep{auer2002finite-time}. Another alternative is Thompson sampling, also termed probability matching \citep{scott2010a-modern}, which selects an arm according to its probability of being optimal according to some underlying probability distribution. For binary (success/failure) rewards, the simple underlying model is to treat $p_k$, the probability of success in arm $k$, as a Beta distribution, which through Bayes rule can be updated based on a sequence of observed successes and failures for each arm. Results using probability matching are highly promising \citep{graepel2010web-scale, granmo2010solving, may2011simulation, chapelle2011an-empirical}, with corresponding theoretical results highlighting the strengths of this approach \citep{agrawal2011analysis, may2012optimistic}.

In many situations, we are not presented with just the arms, but also some covariates $\bm{X}_t$.  As an example, when deciding which ad to display to a user in online advertising, we know which browser and operating system is being used as well as an approximate location based on geolocated IP addresses.  As such, under binary rewards $p_k$ might be a function of these covariates, $p_{tk} = p_k(\bm{X}_t)$.  MABs with covariate information, also called contextual bandits, have been studied in, for example, \citet{yang2002randomized}. \citet{pavlidis2008simulation} show that several commonly-studied empirical approaches, such as greedy, $\epsilon$-greedy, and confidence bound methods can be fit into the contextual situation. Several problems remain with these empirical strategies, however, such as an inability to borrow strength by allowing for hierarchical structure on the covariates.

Some other issues which commonly arise in MAB problems are dynamic expected rewards, also called \textit{restless bandits}, where the $p_k$ may change over time \citep{whittle1988restless}, and arm-acquisition, where new arms can enter the system.  Additionally, other issues can arise including penalties for switching arms, the ability to simultaneously pull multiple arms, or the presence of side observations \citep{mahajan2008multi-armed, caron2012leveraging}.  Further, the MAB problem can be embedded within other methods and algorithms, such as to tune Monte Carlo algorithms \citep{wang2011predictive}. This paper presents a coherent, flexible approach for solving bandit problems in all of these circumstances.  Through sequential Monte Carlo \citep{Doucet2001a}, we are able to achieve this flexibility while maintaining efficient, scalable inference.

\section{The Bayesian Bandit Framework}

Assume we have observations $\bm{y}_t = (y_1, \dots, y_t)$ which are the sequence of observed rewards, and $\bm{a}_t = (a_1, \dots, a_t)$ is the strategy, or arm selected, at each time step. Let $\bm{\beta}$ be a collection of parameters which control such features as the respective rewards for each arm or the impact of covariates; here $p_k$ is a function of of $\bm{\beta}$. Let $f(y_t | a_t=k, \bm{\beta})$ be the likelihood function, or reward distribution, and $\pi(\bm{\beta})$ be the prior distribution. As an example, in the case of binary rewards, if
\begin{align*}
f(y_t | a_t=k, \bm{\beta}) = \beta_{k}^{y_t} (1-\beta_{k})^{1-y_t},
\end{align*}
where $\beta_k$ is the probability of success in arm $k$, and for each $k$ we assign a $Beta(\alpha_0, \alpha_1)$ prior,
\begin{align*}
\pi(\beta_k) \propto \beta_k^{\alpha_0 - 1} (1 - \beta_k)^{\alpha_1-1},
\end{align*}
then the resulting posterior after observing $s$ successes in $n$ trials of arm $k$ is $Beta(s + \alpha_0, n - s + \alpha_1)$.  The fundamental idea behind Thompson sampling is to select a sample from each of these posterior distributions, and pull the arm with the highest-valued sample \citep{thompson1933on-the-likelihood, thompson1935on-the-theory}.

\subsection{A General, Hierarchical Bandit Structure}

As noted by \citet{scott2010a-modern} and others, the Thompson sampling framework need not be constrained to the above Beta-Binomial conjugate setup.  However, as more elaborate $f$ and $\pi$ are chosen, the resulting posteriors will often require approximation methods such as Markov chain Monte Carlo or variational methods. Momentarily ignoring this hurdle, if our data are binary we can use a logit, probit, or other link function within $f$ to connect the probability of success to the model parameters $\bm{\beta}$.  The prior $\pi$ can be used to impose structure on the parameters $\bm{\beta}$, such as inducing dependence, hierarchical structure, or hard and soft constraints through some higher-level parameters $\bm{\tau}$.

This simple Bayesian framework provides a natural solution to address many of the bandit problem variants.  For instance, contextual bandits with binary rewards can be modeled by assuming $f(y_t | a_t=k, \bm{\beta})$ is binomial with probability $(\exp -(\beta_{k0} + \beta_{k1} X_t)+1)^{-1}$, and learning the regression parameters $\bm{\beta} = (\beta_{k0}, \beta_{k1})_{k=1:K}$ akin to Bayesian logistic regression.  More generally, the wealth of knowledge about Bayesian online learning and generalized linear models can be brought in to create a highly flexible class of bandit models. Following the above example, we might suspect that the parameters $\beta_{k1}, k=1, \dots, K$ are related with some overall mean $\nu$, so the prior could be extended through a hierarchy, for example
\begin{align*}
\pi(\beta_{11}, \dots, \beta_{k1}) = \mathcal{N}(\nu, \sigma_{\beta}^2),
\end{align*}
and further hierarchies could be imposed on $\nu, \sigma_{\beta}^2$, or other variables \citep{gelman2006data}.

In general, probability matching selects an arm $k$ according its probability of having the highest expected rewards. In the binary reward case, arm $k$ is selected with probability
\begin{align*}
Pr(p_{tk} > p_{tj}; \;\;\;\; \forall j \neq k).
\end{align*}
In simple cases, one can find these $K$ distributions analytically.  However, beyond these few simple cases, approximations must be made to find these distributions. If we generate multiple samples from $\pi(p_{tk} | \bm{y}_t)$ for each arm $k$, we can compare the sample sets to determine the probability of selecting each arm. The origins of Thompson sampling, however, is to take this idea further, and to generate $1$ sample from each of $\pi(p_{tk} | \bm{y}_t), k=1, \dots, K$.  The largest sample is then selected as the strategy at time $t$.

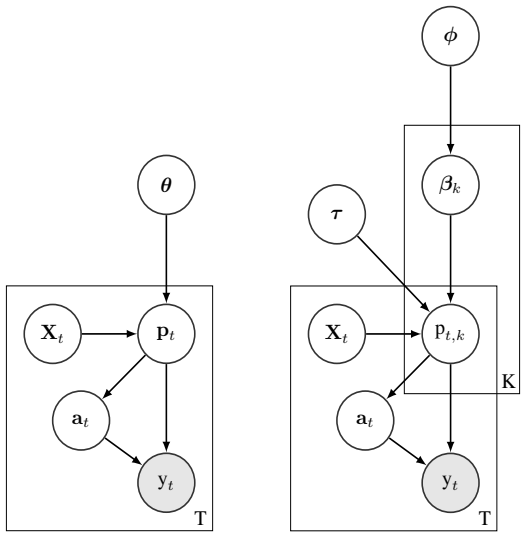
\begin{figure}
\centering
\resizebox{7cm}{7cm}{
\begin{tikzpicture}
\tikzstyle{main}=[circle, minimum size = 10mm, thick, draw =black!80, node distance = 15mm]
\tikzstyle{connect}=[-latex, thick]
\tikzstyle{box}=[rectangle, draw=black!100]
	\node[main, fill = white!100] (theta) {$\bm{\theta}$};
	\node[main] (p) [below=of theta] { $\mathbf{p}_t$ };
	\node[main, fill = black!10] (y) [below=of p] {$\text{y}_t$};
	\node[main] (a) at (-1.5,-4) {$\mathbf{a}_t$};
	\node[main] (x) at (-2, -2.53) {$\mathbf{X}_t$};

	\node[main] (tau) at (3, -.5) {$\bm{\tau}$};

	\node[main] (x2) at (3,-2.53) {$\mathbf{X}_t$};
        \node[main] (p2) at (5, -2.53) {$\text{p}_{t,k}$};
        \node[main,fill = black!10] (y2) [below = of p2]{$\text{y}_t$};
	\node[main] (a2) at (3.5,-4) {$\mathbf{a}_t$};
	\node[main] (beta) [above=of p2] { $\bm{\beta}_k$ };
	\node[main] (phi) [above=of beta] { $\bm{\phi}$ };
	\path (theta) edge [connect] (p)
		(x) edge [connect] (p)
		(p) edge [connect] (a)
		(a) edge [connect] (y)
		(p) edge [connect] (y)
		(phi) edge [connect] (beta)
		(beta) edge [connect] (p2)
		(p2) edge [connect] (y2)
		(p2) edge [connect] (a2)
		(a2) edge [connect] (y2)
		(x2) edge [connect] (p2)

		(tau) edge [connect] (p2);
	\node[rectangle, inner sep=3mm,draw=black!100, fit= (p) (x) (y)] {};
	\node[rectangle, inner sep=0mm,fit= (p) (x) (y), label=below right:T, yshift = -1mm, xshift = -1.5mm] {};

	\node[rectangle, inner sep=5mm,draw=black!100, fit= (beta) (p2), xshift = 2mm] {};
	\node[rectangle, inner sep=0mm, fit= (beta) (p2), label=below right:K, yshift = -13mm, xshift = 2.5mm] {};

        \node[rectangle, inner sep=3mm,draw=black!100, fit= (x2) (y2)] {};
	\node[rectangle, inner sep=0mm, fit= (x2) (y2), label=below right:T, yshift = -1mm, xshift = -1.5mm] {};
\end{tikzpicture}

}
\label{GM:static}
\caption{Graphical model of SMC bandit model.  Left: simple model. Right: with hierarchical parameters $\bm{\phi}$ and across-arm parameters $\bm{\tau}$ separated from arm-specific parameters $\bm{\beta}$.}
\end{figure}

We are interested in learning about the underlying parameters $\bm{\theta} = (\bm{\beta}, \bm{\tau}, \bm{\phi})$ as data is collected, specifically through the posterior $\pi(\bm{\theta} | \bm{y}_t, \bm{X}_t, \bm{a}_t)$.  Consistent with earlier discussion, $\bm{\beta}$ is the set of arm-specific parameters, which are related through some hierarchical parameters $\bm{\phi}$, while $\bm{\tau}$ are parameters shared across arms.  Figure \ref{GM:static} displays the model with and without the parameter space separated into components.  Several bandit problem extensions can be fit into this model. For example, arms being added and removed are automatically included in the model; in practice, one need only expand the parameter vector to include the additional parameters.  Further, costs for switching arms may be included by directly modeling a penalty parameter which reduces the expected reward of all arms other than the most recently chosen.

\section{Efficient Inference}
 
In restricted conjugate cases, the above model could be solved in closed form.  However, more generally some approximation of $\pi(\bm{\theta} | \bm{y}_t, \bm{X}_t, \bm{a}_t)$ is required for each $t$.  A simple, if computationally expensive, approach is to draw samples using MCMC or related methods for each $t$.  While such an approach is intuitively straightforward, in practice the computational cost of sampling from $\pi(\bm{\theta} | \bm{y}_t, \bm{X}_t, \bm{a}_t)$ increases with $t$.  As an example, in the probit regression case Gibbs sampling is possible through sampling of a latent variable for each binary $y_t$ \citep{Albert1993a}.  As a more efficient alternative, we propose to use sequential Monte Carlo (SMC) methods to transition through the sequence of distributions $\{\pi(\bm{\theta} | \bm{y}_t, \bm{X}_t, \bm{a}_t)\}_{t=1,\dots,T}$ in an efficient manner \citep{Doucet2001a}. Intuitively, the approximation at time $t$ is leveraged to quickly and efficiently make an approximation at time $t+1$.  

SMC methods were originally designed for sequences of distributions of increasing dimension, with applications to state-space models \citep{doucet2000on-sequential} and target tracking \citep{Liu1998a} among others, though they have recently been shown to provide flexibility and efficiency in static-dimensional problems as well \citep{Chopin2002a, Del-Moral2006a, Bornn2010b}.  The goal of SMC methods is to sample from this sequence of distributions sequentially; because SMC borrows information from adjacent distributions, it will typically be computationally cheaper than MCMC even if we can sample from each distribution using MCMC.

For each time $t$, SMC collects $N$ weighted samples (often called particles) $\{\bm{w}_t^{(i)}, \bm{\theta}_t^{(i)}\}$, $i=1,...,N$ approximating $\pi_t = \pi(\bm{\theta} | \bm{y}_t, \bm{X}_t, \bm{a}_t)$.  Expectations with respect to this posterior may be calculated with the weighted samples using $\hat{E}_{\pi_t}(g(\bm{\theta})) = \sum_{i=1}^N \bm{w}_{t}^{(i)} \cdot g(\bm{\theta}_{t}^{(i)})$.  Through importance sampling, these (weighted) particles can then be reweighted to approximate the subsequent distribution.  To ensure the sample does not become degenerate, the effective sample size (ESS),
\begin{align*}
\frac{1}{\sum_{i=1}^N (W_t^{(i)})^2},
\end{align*}
is monitored and the particles are resampled when the ESS drops below some threshold $c$; often $c=N/2$.  While particle filters are often criticized for problems of degeneracy, our interest is in each individual distribution rather than the entire joint distribution over the whole sample path.  As such, resampling and related tools are available to straightforwardly mitigate this issue \citep{Doucet2001a}.

SMC methods have been employed in sequential decision-making problems previously. Specifically, \citet{Yi:2009fk} used SMC to emulate human performance on beta-binomial restless bandit problems.  In another related area, \citet{Coquelin:2008uq} explored a partially observable Markov decision problem (POMDP), where SMC was used alongside a policy gradient approach to optimize the decision at each point in time.

In the dynamic case discussed later, diversity is naturally incorporated into the samples through the prior dynamics.  However, in the static case, we also move the particles with a Markov kernel of invariant distribution $\pi_t$ when we resample.  In cases where the kernel is known to mix slowly, one may wish to move the particles multiple times at each resampling step.  Of course, in these situations MCMC will similarly suffer due to the slow mixing of the kernel. Algorithm \ref{alg:static} shows the SMC bandits algorithm for static parameters:
\begin{algorithm}
\caption{Sequential Monte Carlo bandits}
\label{alg1}
\begin{algorithmic}
\REQUIRE N
\STATE Obtain $N$ samples $\bm{\theta}_0^{(i)}, i=1, \dots, N$,  from $\pi(\bm{\theta})$
\STATE $\bm{w}_0^{(i)} \leftarrow N^{-1}, i=1, \dots, N$
\FOR{$t=1,\dots,T$}
	\STATE Select particle $i'$ according to probabilities $\bm{w}_t$
	\STATE $a_t \leftarrow \max_k(p_k(\bm{\theta}_{t-1}^{(i')}, \bm{X}_t))$
	\STATE Observe $y_t$
	\STATE $\bm{w}_t \leftarrow \bm{w}_{t-1} f(y_t | \bm{\theta}_{t-1})$
	\STATE ESS $\leftarrow (\sum_{i=1}^N (\bm{w}_t^{i})^2)^{-1}$
		\IF{ESS $> c$}
			\STATE $\bm{\theta}_{t} \leftarrow \bm{\theta}_{t-1}$
		\ELSE
			\STATE Resample $\bm{\theta}_{t}$ with probabilities $\bm{w}_t$
			\STATE $\bm{w}_t^{i} \leftarrow N^{-1}, i=1, \dots, N$
			\STATE Move $\bm{\theta}_t$ with $K( \cdot | \theta_t)$, a kernel with invariant distribution $\pi(\bm{\theta} | \bm{y}_t, \bm{X}_t, \bm{a}_t)$
		\ENDIF
\ENDFOR
\end{algorithmic}
\label{alg:static}
\end{algorithm}
SMC bandits has, at its core, a vanilla SMC algorithm.  Based on the large literature on SMC methods, more advanced alternatives exist.  For example, sampling from the prior may be inefficient, in which case we can design more effective proposals, as in \citet{Van-der-Merwe2001a}. An important side benefit from this algorithm is that we have, automatically, samples from the posterior distribution $\pi_t = \pi(\bm{\theta} | \bm{y}_t, \bm{X}_t, \bm{a}_t)$ at every iteration.  As a result, Thompson sampling amounts to randomly selecting one of the particles according to the weights $\bm{w}_t$, and finding the arm with highest expected reward according to the value of that particle.  Using all of the particles, one could also perform probability matching through a Monte Carlo estimate of $Pr(p_{tk} > p_{tj}; j=1, k-1, k+1, K)$.

\subsection{Contextual Bandit Comparison}

While sequential Monte Carlo bandits (hereafter SMC bandits) will naturally accommodate hierarchical model structure, dependence between parameters, and other structure, existing approaches for bandit problems do not naturally handle such issues.  As such, we choose to compare SMC bandits in an arena where alternatives, such as $\epsilon$-greedy and upper confidence bound (UCB) methods, are naturally suited, namely vanilla contextual bandits.  Additionally, due to the additional parametric assumptions inherent in SMC bandits, we use diffuse prior information to simulate real-world modeling ignorance.  The simulation data-generating mechanism is as follows:
\begin{align*}
	\beta_{k0} &\sim U(-1,1)\\
	\beta_{k1} &\sim N(0,1)\\
	\beta_{k2} &= 1\\
	\bm{X}_{t1},\bm{X}_{t2} &\sim N(0,1)\\
	y_t | a_t=k, \bm{X}_t, \bm{\beta} &\sim Bern(\Phi(\beta_{k0} + \beta_{k1}\bm{X}_{t1} + \beta_{k2}\bm{X}_{t2} ))
\end{align*}
where $\Phi$ is the normal CDF and we have $k=1, \dots, 4$ arms.  We simulate data of length $t=1, \dots, T$ where $T=10,000$, and take averages over 50 repeated simulations of the above system.  For SMC bandits, we use the same prior for all elements of $\bm{\beta}$, namely $N(0,10)$.  This diffuse prior covers the generated data, but has roughly $10$ times the variance, representing apriori uncertainty about the model parameters.

We compare SMC bandits to contextual $\epsilon$-Greedy and a UCB methods \citep{li2010a-contextual-bandit,filippi2010parametric}.  For both we use the same probit link as in the data-generation mechanism.  For $\epsilon$-greedy we try both $\epsilon=0.1$ and $\epsilon=0.01$, and for UCB we explore $90\%$ and $95\%$ confidence intervals.

\begin{figure}
\centering
\includegraphics[width=220px]{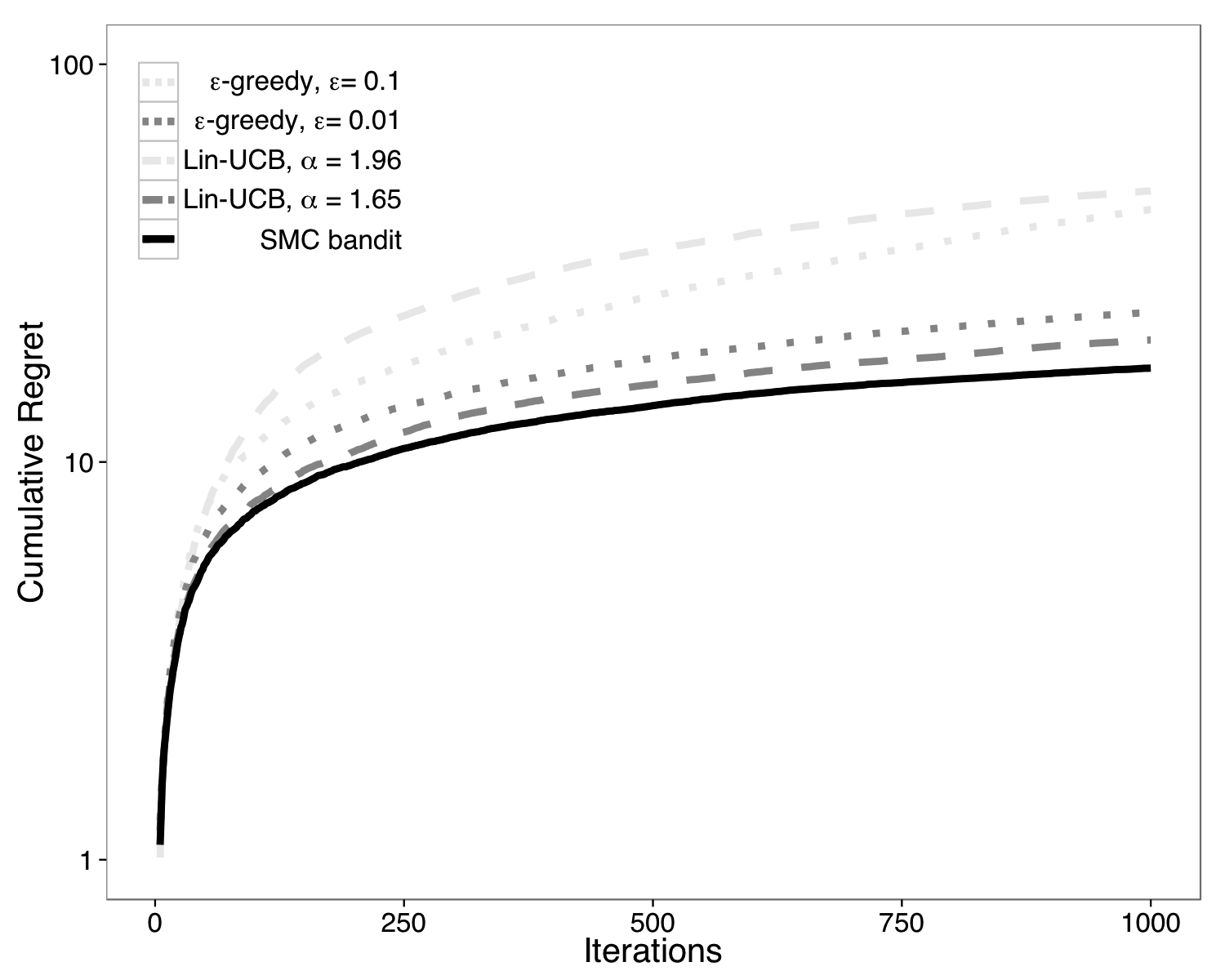}
\caption{Cumulative regret for contextual bandits for $\epsilon$-greedy vs. UCB vs. SMC bandits}
\label{regret}
\end{figure}

Given the covariate set, we know the ground truth optimal strategy at each time step.  As such, we compare the cumulative regret from each method, which is the cumulative difference between strategies of the methods vs. the optimal strategy.  Figure \ref{regret} shows the cumulative regret, where we notice that SMC bandits have the lowest cumulative regret at each time point, even in this simple problem.  In contrast, the $\epsilon$-Greed and UCB methods' performance is quite sensitive to the selection of $\epsilon$ or the confidence level, respectively. It is worth reiterating that the SMC bandits framework allows one to model hierarchies, constraints, and other structure naturally, while the other methods do not naturally extend to these cases.

\begin{figure}
\centering
\includegraphics[width=220px]{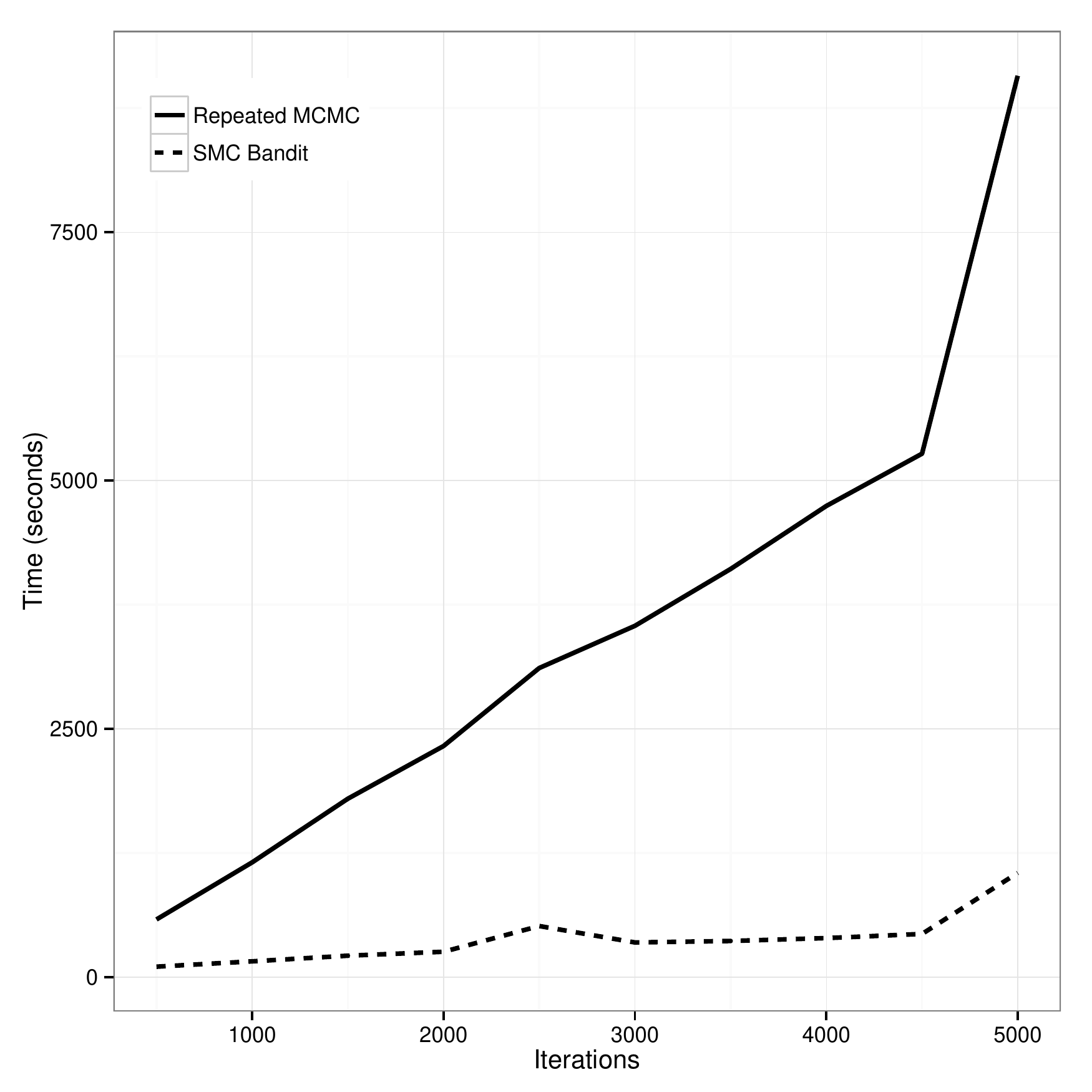}
\caption{Time comparison of SMC bandits vs. using repeated MCMC for $T=500,1000,\dots,5000$.  Number of samples for each method is $1000$.}
\label{MCMC_vs_SMC}
\end{figure}

Additionally, we also compare the use of SMC in the Bayesian hierarchical bandit framework to instead using MCMC repeated at each iteration.  Figure \ref{MCMC_vs_SMC} shows the time taken for each method, using $1000$ samples for each.  We see that SMC provides a significant reduction in computational time in comparison to repeating MCMC at each iteration.  The intuition is that due to the efficient reweighting mechanism of the SMC algorithm, it only occasionally needs to move samples with a Markov kernel, and as such saves considerable computational cost. The improvements are similar to those from using SMC rather than MCMC for the tasks of prior sensitivity analyis and cross-validation \citep{Bornn2010b}.

\section{Contextual Dynamic Bandits}

The case where the reward is changing over time is often referred to as restless, or dynamic, bandits.  Intuitively, the difficulty arises in that if a given action is not taken for a period of time, it is possible that the unobserved action has had an increase in its expected reward; as a result, strategies for restless bandits will generally not converge to a single action, as on some regular basis other arms must be ``checked'' to monitor for changes in expected reward.

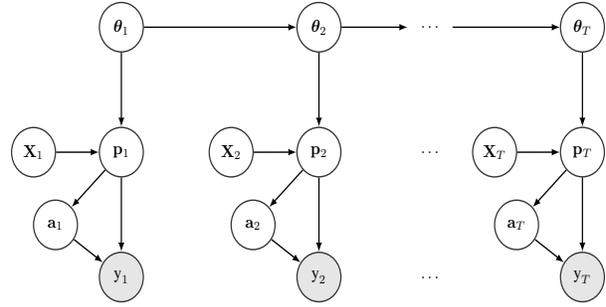
\begin{figure}
\centering
\resizebox{8cm}{4cm}{
\begin{tikzpicture}
\tikzstyle{main}=[circle, minimum size = 10mm, thick, draw =black!80, node distance = 15mm]
\tikzstyle{connect}=[-latex, thick]
\tikzstyle{box}=[rectangle, draw=black!100]
	\node[main, fill = white!100] (theta) {$\bm{\theta}_1$};
	\node[main] (p) [below=of theta] { $\mathbf{p}_1$ };
	\node[main, fill = black!10] (y) [below=of p] {$\text{y}_1$};
	\node[main] (a) at (-1.5,-4) {$\mathbf{a}_1$};
	\node[main] (x) at (-2, -2.53) {$\mathbf{X}_1$};

	\node[main, fill = white!100] (theta2) at (4.5, 0) {$\bm{\theta}_2$};
	\node[main] (p2) [below=of theta2] { $\mathbf{p}_2$ };
	\node[main, fill = black!10] (y2) [below=of p2] {$\text{y}_2$};
	\node[main] (a2) at (3,-4) {$\mathbf{a}_2$};
	\node[main] (x2) at (2.5, -2.53) {$\mathbf{X}_2$};

	\node[main, draw = none] (invisible) [right = of theta2] {$\mathbf{\dots}$};
	\node[main, draw = none] (invisible2) [below = of invisible] {$\mathbf{\dots}$};
	\node[main, draw = none] (invisible3) [below = of invisible2] {$\mathbf{\dots}$};

	\node[main, fill = white!100] (theta3) at (10.5, 0) {$\bm{\theta}_T$};
	\node[main] (p3) [below=of theta3] { $\mathbf{p}_T$ };
	\node[main, fill = black!10] (y3) [below=of p3] {$\text{y}_T$};
	\node[main] (a3) at (9,-4) {$\mathbf{a}_T$};
	\node[main] (x3) at (8.5, -2.53) {$\mathbf{X}_T$};

	\path (theta) edge [connect] (p)
		(x) edge [connect] (p)
		(p) edge [connect] (a)
		(a) edge [connect] (y)
		(p) edge [connect] (y)
		
		(theta) edge [connect] (theta2)
		(theta2) edge [connect] (p2)
		(x2) edge [connect] (p2)
		(p2) edge [connect] (a2)
		(a2) edge [connect] (y2)
		(p2) edge [connect] (y2)

		(theta2) edge [connect] (invisible)

		(invisible) edge [connect] (theta3)
		(theta3) edge [connect] (p3)
		(x3) edge [connect] (p3)
		(p3) edge [connect] (a3)
		(a3) edge [connect] (y3)
		(p3) edge [connect] (y3);
		



\end{tikzpicture}
}
\label{GM_dynamic}
\caption{Graphical model of SMC bandits for dynamic rewards.  As with Figure 1, the hierarchical and shared parameters could also be separated.}
\end{figure}

The framework proposed here provides a natural representation of restless bandits as a hidden Markov model (Figure \ref{GM_dynamic}), where the hidden states $\bm{\theta}$ which determine the expected reward vary in time \citep{gupta2011thompson}.  We notate these time-varying parameters as $\bm{\theta}_t$.

\begin{algorithm}
\caption{Dynamic reward sequential Monte Carlo bandits}
\label{alg1}
\begin{algorithmic}
\REQUIRE N
\STATE Obtain $N$ weighted samples $\bm{\theta}_0^{(i)}$ from $\pi(\bm{\theta})$
\STATE $\bm{w}_0^{i} \leftarrow N^{-1}, i=1, \dots, N$
\FOR{$t=1,\dots,T$}
	\STATE Sample $\bm{\theta}_t$ from $\pi(\cdot|\bm{\theta}_{t-1})$
	\STATE Select particle $i'$ according to probabilities $\bm{w}_t$
	\STATE $a_t \leftarrow \max_k(p_i(\bm{\theta}_t^{(i')}, X_t))$
	\STATE Observe $y_t$
	\STATE $\bm{w}_t \leftarrow \bm{w}_{t-1} f(y_t | \bm{\theta}_{t})$
	\STATE ESS $\leftarrow (\sum_{i=1}^N (\bm{w}_t^{i})^2)^{-1}$
		\IF{ESS $< c$}
			\STATE Resample $\bm{\theta}_{t}$ with probabilities $\bm{w}_t$
			\STATE $\bm{w}_t^{i} \leftarrow N^{-1}, i=1, \dots, N$
		\ENDIF
\ENDFOR
\end{algorithmic}
\end{algorithm}
Due to their probabilistic nature, SMC bandits will not only estimate the evolving parameters $\bm{\theta}_t$, but also their corresponding uncertainty.  As such, when a given arm is not sampled for a period of time, its estimated uncertainty grows; more specifically, its posterior variance increases, and as a result over time that arm will be re-selected due to the Thompson sampling mechanism.  Numerically, the prior dynamics increase the variance of the particles for the unsampled arm; as such, when the Thompson sampling mechanism selects a given particle, it becomes increasingly more likely that that arm has the largest value.  We demonstrate this with a simulated example.

\subsection{Dynamic Bandit Comparison}

\begin{figure}
\centering
\includegraphics[width=240px, height=240px]{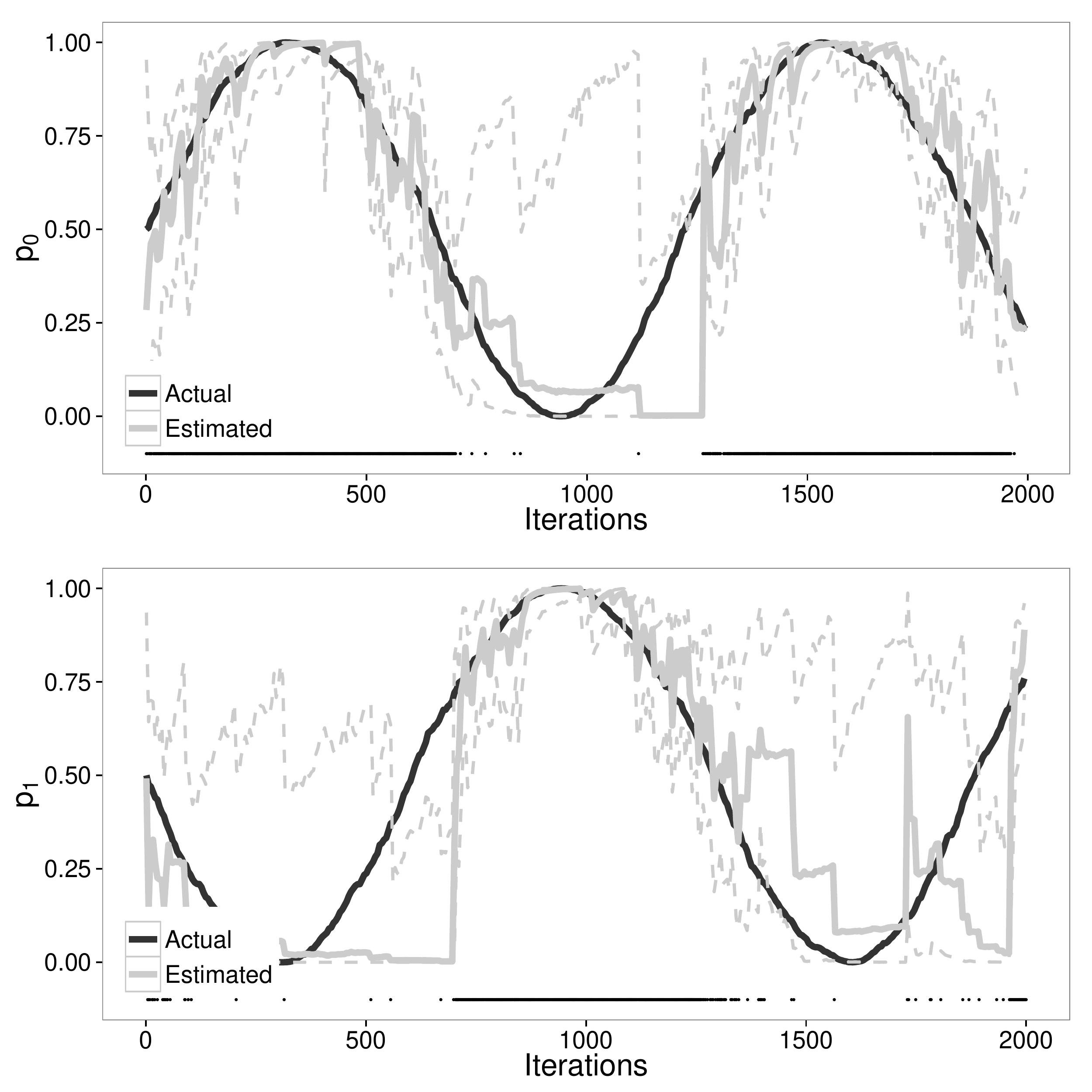}
\caption{Tracking of dynamic rewards with associated uncertainties from dynamic SMC bandits.  Top: Arm 1.  Bottom: Arm 2.  The true parameters are shown, as well as the estimated posterior mean and variance. We see that when an arm has higher parameter value (and hence higher reward) it is more likely to be sampled, and when it is not sampled, its variance grows.}
\label{dynamic}
\end{figure}

\begin{figure}
\centering
\includegraphics[width=220px, height=160px]{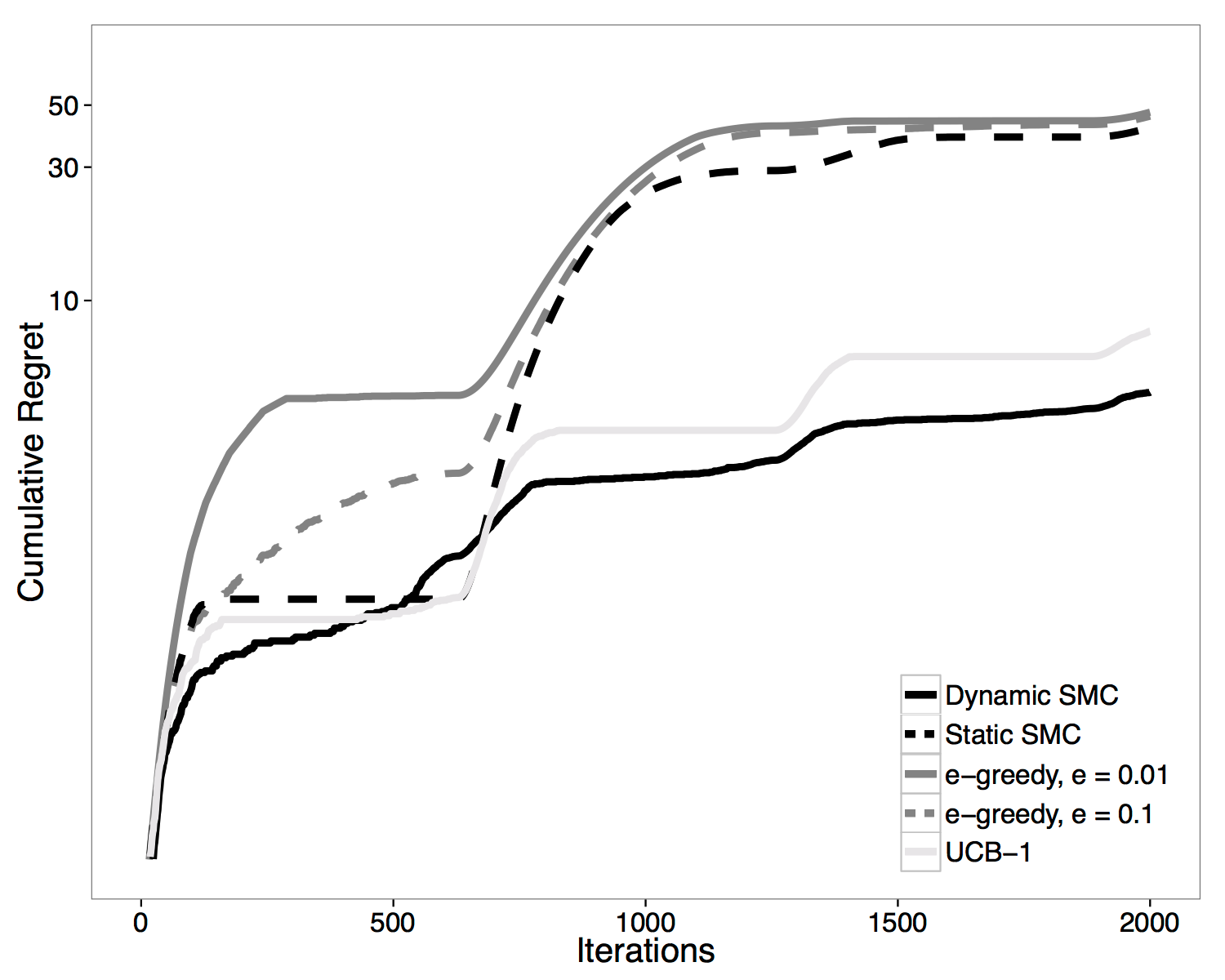}
\caption{Cumulative regret of various MAB algorithms on dynamic reward simulation.  Dynamic SMC dominates the other methods, which can be seen to have a jump in regret when the highest-reward policy switches from arm 1 to arm 2.}
\label{dynamic_regret}
\end{figure}

We study the ability of SMC bandits to track dynamic rewards, simulating data in the following manner
\begin{align*}
	\beta_{0,t} &= \Phi^{-1}(\frac{1}{2} [ \sin(t/100) + 1] ) \\
	\beta_{1,t} &= \Phi^{-1}(\frac{1}{2} [ \sin(t/100 + \pi) + 1] ) \\
	y_t | a_t=k, \bm{\beta} &\sim Bern(\Phi(\beta_{k,t})), \; t=1\dots2000
\end{align*}
where we have $k=1, 2$ arms. We impose the following model structure for the dynamic SMC bandit:
\begin{align*}
	\beta_{k,0} &\sim N(0,1)\\
	\beta_{k,t} &\sim N(\beta_{k,t-1},1)\\
	y_t | a_t=k, \bm{\beta} &\sim Bern(\Phi(\beta_{k,t})).
\end{align*}
Figure \ref{dynamic} shows the tracking of the instantaneous probabilities $\Phi(\beta_{k,t})$, which control the dynamic expected reward at time $t$.  We notice that the arm with highest expected reward (as indicated by larger parameter values) is sampled more often, and posterior variance of the unsampled arm grows until it is next chosen. In contrast, the optimal arm is sampled more frequently, and hence the posterior mean better tracks the true reward.  In Figure \ref{dynamic_regret}, we plot the cumulative regret for the dynamic SMC method, as well as several other algorithms. We see that the dynamic SMC method has lowest regret; the figure also shows that as the optimal arm changes around iteration $700$, the regret of all the methods increases, with the dynamic SMC method adapting to the change and hence showing the least increase in regret.

Interestingly, and somewhat paradoxically, the dynamic SMC bandit method is generally faster than the static case due to the prior dynamics inducing diversity, rather than the static SMC bandit's use of a Markov step.  While one might be tempted to model static problems in the dynamic framework, this is equivalent to assuming your (static) parameters vary stochastically.  As a result, arms which you establish with high certainty to be low-reward early on will come back into play later in the sequence due to the artificial dynamics induced on the parameters.  As a result, the computational gains will come with the cost of incurring additional regret.

\section{Empirical Results}

An ideal performance evaluation of a bandit algorithm would include testing the algorithm online on a real system. Because online testing is expensive and typically unfeasable in most situations, we evaluate the empirical performance of the SMC bandit offline.

More precisely, we assume that our empirical data consists of a sequence of arm selections $\bm{a}_t = (a_1, \dots, a_t)$, as well as the reward $\bm{y}_t = (y_1, \dots, y_t)$ and optional contexts $\bm{X}_t = (\bm{X}_1, \dots, \bm{X}_t)$. Crucially, the reward $y_t$ is only observed for the arm $a_t$ that was chosen uniformly at random, at time $t$. 

\begin{algorithm}
\caption{Empirical Bandit Performance Evaluator}
\label{unbiased_bandit}
\begin{algorithmic}
\REQUIRE bandit algorithm $\mathcal{B}$; data $\mathcal{D}_T = (\bm{X}_T, \bm{a}_T, \bm{y}_T)$
\STATE $h_0 \leftarrow \emptyset$ (Initially empty history)
\STATE $R_\mathcal{B} \leftarrow 0$ (Initially zero total reward for $\mathcal{B}$)
\FOR{$t=1,\dots,T$}
	\IF{$\mathcal{B}(h_{t-1}, \bm{X}_t) = a_t$}
		\STATE $h_t \leftarrow$  concatenate$(h_{t-1}, (\bm{X}_t, a_t, y_t))$
		\STATE $R_\mathcal{B} \leftarrow R_\mathcal{B} + y_t$
	\ELSE
		\STATE continue
	\ENDIF
\ENDFOR
\end{algorithmic}
\end{algorithm}

The proposed offline bandit evaluator is shown in Algorithm \ref{unbiased_bandit} \citep{LiUnbiased2011}. The method requires bandit algorithm $\mathcal{B}$ as well as empirical data $\mathcal{D}$, containing $T$ realizations of the bandit data $\{(\bm{X}_t, a_t, y_t)\}_{t = 1}^{T}$. At each time $t$, we keep a history $h_t$, which contains a subset of the overall data $\mathcal{D}$ presented to the bandit algorithm. At time $t = 0$, $h_0$ is empty. For each sample in $\mathcal{D}$, we ask the bandit algorithm for a recommended arm to play. If, given the current history $h_{t-1}$, the algorithm $\mathcal{B}$ selects the same arm as the arm $a_t$ described by the data, then the data sample is retained (added to the history) and the total reward earned by the bandit algorithm $R_\mathcal{B}$ is updated. Otherwise, if $\mathcal{B}$ selects a different arm, the data sample is ignored and the bandit algorithm proceeds to the next data sample without any change in its state. 

We assume that the original data contains arms that are chosen uniformly at random, so each data sample is retained by the bandit algorithm $\mathcal{B}$ with probability $\displaystyle \frac{1}{K}$. Therefore the retained samples have the same distribution as the original data set $\mathcal{D}$ and the evaluator in Algorithm \ref{unbiased_bandit} is functionally equivalent to (an unbiased estimator of) an online evaluation of $\mathcal{B}$ \citep{LiUnbiased2011}. 

\subsection{Online Advertising Recommendation}

Using the offline evaluation technique described in Section 6.1, we evaluate the performance of our SMC bandit algorithm on real-world advertising data. The data is sourced from an online video advertising campaign in 2011, which marketed an online dating service for older singles. 

In particular, the dataset contains 93,264 impressions, where an impression describes the event of displaying an advertisement to a particular user. There exists a set of four advertisements from which an advertisement can be shown to the user. Each impression consists of the advertisement shown to the user, whether the user clicked on the advertisement, and relevant contextual information, including geo-location (latitude and longitude of the user's location in the United States) and datetime (time of day and day of week the advertisement was shown). 

We formulate this data into a 4-armed bandit problem by modeling the advertisements as arms, the clicks as binary rewards, and adopting a probit link function to map the linear function of our contextual information to the binary reward distribution, a.k.a. the probit SMC bandit. We test the performance of two variations of the probit SMC bandit, the static and dynamic case, using the offline evaluation technique described in Section 6.1. Each SMC algorithm is evaluated 100 times (to reduce variance of our performance estimate) over all 665,321 impressions, initialized with $N = 1000$ particles and ESS threshold set to $500 = N/2$. Similarly, we test the performance of the contextual $\epsilon$-greedy and UCB algorithms, as well as a random arm selector, as a baseline comparison. Results are shown in Table 1.

\begin{table}[h]
\centering
    \begin{tabular}{|l|c|c|c|}
        \hline
        Bandit                        & Reward  & \% diff & p-value\\ \hline
        Random (Baseline)                                  & 0.0079                    & -                &   -    \\ 
	$\epsilon$-greedy, $\epsilon=0.1$ & 0.0082 & +3.7 & 0.47 \\
	$\epsilon$-greedy, $\epsilon=0.01$ & 0.0078 & -1.2 & 0.65 \\
	UCB-1 & 0.0078 & -1.2 & 0.78 \\
        Static SMC                        & 0.0077                  & -2.5                &    0.55    \\ 
        Dynamic SMC                       & 0.0088                  & + 11.4         &   $<$ 0.01          \\
        \hline
    \end{tabular}
\caption{Average reward per iteration earned from the application of various 4-armed bandit algorithms on an online advertising dataset. Column 3 is computed as the percentage difference of the average reward as compared to the random metric (baseline). Column 4 reports p-values using Student's $t$-test for the difference in means.}
\end{table}

We notice that, out of the various bandit algorithms tested, the dynamic SMC bandit performs the best, at about a $10\%$ improvement over the baseline random arm selector. Interestingly, while the performance of the dynamic SMC bandit is the best out of the group of algorithms, the static SMC bandit has middling performance. This is a strong indication that the reward distributions for each of the advertisements in the online advertising dataset move dynamically, which makes intuitive sense, because customer preferences for different advertisements likely vary over time. Nonstationary reward distributions prove to be a problem for the MCMC step in the static SMC bandit, which may be the cause of the decreased average reward. In fact, we also see that the lin-$\epsilon$-greedy algorithm with $\epsilon = 0.1$ performs second best - another indication that the bandit arms may be moving dynamically. Recall that higher values of $\epsilon$ in $\epsilon$-greedy algorithms force the algorithm to perform more random exploration. A bandit algorithm that performs random exploration at a high rate is particularly suited for dynamic rewards, as the algorithm can better track the random distributional movements of the arms it explores. This is likely why the $\epsilon = 0.1$ greedy algorithm slightly outperforms the $\epsilon = 0.01$ greedy algorithm. 
\begin{figure}
\centering
\includegraphics[width=240px, height=240px]{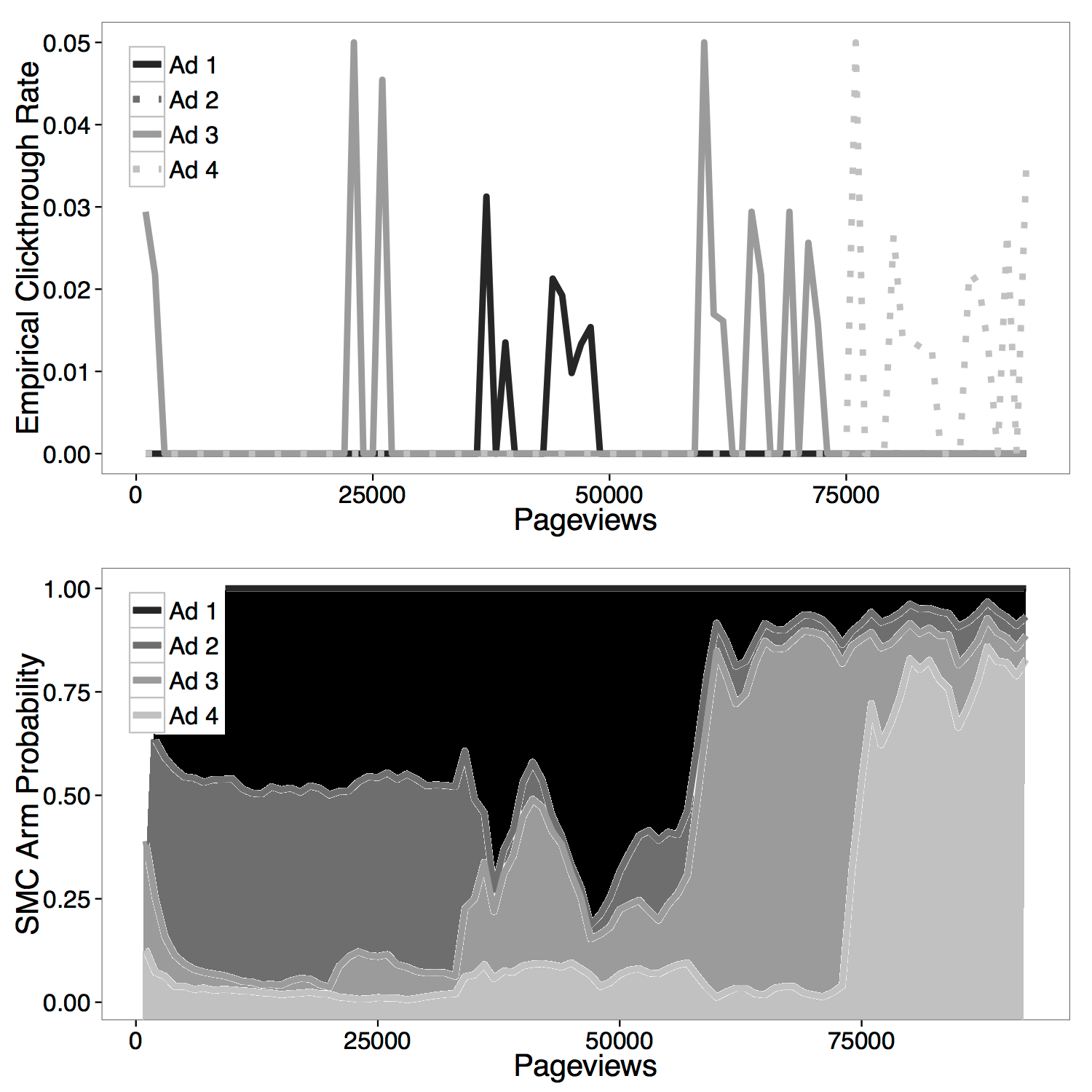}
\caption{Comparison of empirical clickthrough rates for online advertising dataset and dynamic SMC arm (advertisement) selection probabilities. Top: Empirical clickthrough rates. Bottom: SMC arm selection probabilities as stacked area graph. Note that as empirical clickthrough rate increases, so does the corresponding area in the bottom are graph. Results were obtained by binning the 93,264 impressions into bins of 1,000 impressions and calculating the clickthrough rates for each bin. }
\label{moving_average}
\end{figure}

A simple binning of the overall clickthrough rate of each advertisement in our dataset confirms our suspicion that the arm reward distributions are moving dynamically (Figure \ref{moving_average}). Figure \ref{moving_average} shows the SMC bandit estimated probability of displaying each arm at every pageview, using the evaluation method described in Algorithm \ref{unbiased_bandit}. Note the dynamic nature of the empirical clickthrough rates, as well as the similarity of the SMC estimated means to the empirical rates. This analysis provides conclusive evidence of the dynamic nature of the advertising data, as well as the efficacy and robustness of the SMC algorithm, especially in the dynamic case. While the other popular bandit algorithms exhibiti similar performance to random arm selection, the dynamic SMC bandit show significant improvement over the other arm allocation methods. 

These results provide promising indications of the applicability of the SMC bandit in real-world situations. Especially in an application like online advertising, in which millions of advertisements are shown to users every day, a $10\%$ improvement in the average clicks per user can potentially lead to an extraordinary increase in revenue for advertisers.

\section{Conclusions}

In conclusion, we have proposed a flexible and scalable, yet simple, method for learning in multi-armed bandit problems.  The proposed framework naturally handles contexts, dynamic rewards, the additional and removal of arms, and other features common to bandit problems.  Through a hierarchical Bayesian model, the method is highly adaptable in its specification, with the user able to adjust hierarchical structure, hard and soft constraints, and other features through the prior specification of the parameter set.  In addition, this additional structure does not come at a significant cost, as the sequential Monte Carlo algorithm proposed allows for scalability in time, but also as model complexity increases.

While using Bayesian inferential methods for modeling multi-armed bandit problems is not new, the hierarchical and dynamic structure proposed here provides for significantly increased flexibility. The framework allows for borrowing of the strength of the large literature on hierarchical Bayesian methods as well as the literature on sequential Monte Carlo to allow natural extendability.


\bibliographystyle{apalike}
\bibliography{lukebornn_SMCBandits}

\end{document}